# Psychomatics - A Multidisciplinary Framework for Understanding Artificial Minds

Giuseppe Riva [1-2], Fabrizia Mantovani [3], Brenda K. Wiederhold, Ph.D. [4-5]
Antonella Marchetti [6-7], Andrea Gaggioli [2-8]

[1] Humane Technology Lab., Catholic University of Sacred Heart, Milan, Italy
[2] Applied Technology for Neuro-Psychology Lab., Istituto Auxologico Italiano IRCCS, Milan, Italy
[3] Centre for Studies in Communication Sciences "Luigi Anolli" (CESCOM), Department of Human Sciences for Education ''Riccardo Massa'', University of Milano - Bicocca, Milan, Italy
[4] Virtual Reality Medical Center, La Jolla, CA, USA
[5] Interactive Media Institute, San Diego, CA, USA
[6] Department of Psychology, Catholic University of Sacred Heart, Milan, Italy
[7] Research Unit on Theory of Mind, Catholic University of Sacred Heart, Milan, Italy
[8] Research Center in Communication Psychology (PSICOM), Catholic University of Sacred Heart, Milan, Italy

**Abstract**: Although LLMs and other artificial intelligence systems demonstrate cognitive skills similar to humans, like concept learning and language acquisition, the way they process information fundamentally differs from biological cognition. To better understand these differences this paper introduces Psychomatics, a multidisciplinary framework bridging cognitive science, linguistics, and computer science. It aims to delve deeper into the high-level functioning of LLMs, focusing specifically on how LLMs acquire, learn, remember, and use information to produce their outputs. To achieve this goal, Psychomatics will rely on a comparative methodology, starting from a theory-driven research question - is the process of language development and use different in humans and LLMs? - drawing parallels between LLMs and biological systems.

Our analysis shows how LLMs can map and manipulate complex linguistic patterns in their training data. Moreover, LLMs can follow Grice's Cooperative Principle to provide relevant and informative responses. However, human cognition draws from multiple sources of meaning, including experiential, emotional, and imaginative facets, which transcend mere language processing and are rooted in our social and developmental trajectories. Moreover, current LLMs lack physical embodiment, reducing their ability to make sense of the intricate interplay between perception, action, and cognition that shapes human understanding and expression.

Ultimately, Psychomatics holds the potential to yield transformative insights into the nature of language, cognition, and intelligence, both artificial and biological. Moreover, by drawing parallels between LLMs and human cognitive processes, Psychomatics can inform the development of more robust and human-like AI systems.

Please send correspondence to:

Prof. Giuseppe Riva
Università Cattolica del Sacro Cuore
Largo Gemelli 1
20123 Milan, Italy
Email: giuseppe.riva@unicatt.it
Phone: +39-02-72343734

# Psychomatics - A Multidisciplinary Framework for Understanding Artificial Minds

## 1. Introduction: Defining Psychomatics

In the rapidly evolving landscape of artificial intelligence (AI) [1-4], there exists an ongoing debate about how we should evaluate and understand the cognitive capacities of AI systems [5,6]. Traditional approaches have often centered on either using automated evaluations based on specific benchmarks, or comparing AI behaviors with those of humans, aiming to identify similarities and differences [7-9].

For example, Flan-PaLM, an instruction-tuned variant of the Pathways Language Model (PaLM) achieved 67.6% accuracy on MedQA, the US Medical Licensing Exam-style questions [10]. Another study [11] showed that a Large Linguistic Model (LLM) achieved human-like performances in a set of compositional tasks, demonstrating systematic compositionality—the algebraic ability to understand and produce novel combinations from known components. Finally, Hagendorff, et al. [12] found that human-like intuitive behavior and reasoning biases emerged in large language models but disappeared in ChatGPT, This LLM responded correctly to a great majority of tasks, decisively outperforming humans in their ability to avoid traps embedded in the tasks.

Several other studies have been conducted to compare the behavior of AI and humans, identifying both similarities and differences. For example, Mei et al. [13] used conventional behavioral measures from economics and psychology to evaluate the behavioral and personality traits of large language models (LLMs). The study found that ChatGPT-4 exhibited traits that were indistinguishable from those of a random human among tens of thousands of human participants. Based on these findings, Meng [14] signals the emergence of a new research area called "AI Behavioral Science" (AIBS), which uses methods from human behavioral science to examine and design AI behavior.

However, this emerging field does not provide a deeper insight into AI [15], although it does demonstrate that AI has advanced information manipulation capabilities. As noted recently by Heaven [16]: "The largest models, and large language models in particular, seem to behave in ways textbook math says they shouldn't. This highlights a remarkable fact about deep learning, the fundamental technology behind today's AI boom: for all its runaway success, nobody knows exactly how—or why—it works."

We argue that this shortcoming arises because the AIBS overlooks two critical aspects. First, it fails to clarify how the ability of LLMs to use information to generate outcomes compares to or differs from that of humans. [17]. LLMs' abilities may differ from human processes. If so, using LLMs as human subjects in psychology experiments is useless. Secondly, like behaviorism in psychology and the imitation test proposed by Alan Turing, AIBS concentrates only on the observable aspects of behavior and their environmental determinants. While LLMs can adeptly pass the Turing test by emulating human-like responses [18], the adage "doing the same thing for different reasons" underscores that merely replicating external behaviors is inadequate for providing a coherent and comprehensive account of the emergence and functioning of cognitive abilities [19]. Without exploring the intrinsic processes that underlie these abilities, we risk falling into the trap of "conceptual borrowing" [20], where we anthropomorphize AI systems as computational analogues of the human brain and ascribe psychological properties to them without a firm theoretical or empirical foundation [21].

In this Perspective, we propose a multidisciplinary synergy among cognitive science, linguistics, and computer science to uncover groundbreaking insights into large language models (LLMs). We present "Psychomatics," (the crasis of the terms "psychology" and "informatics") an interdisciplinary framework that connects these fields, focusing on how artificial intelligence (AI), and in particular LLMs, processes information. This framework explores how LLMs perceive, learn, remember, and employ information to generate outputs. To achieve this goal, Psychomatics will rely on a comparative methodology, drawing parallels between LLMs and biological systems.

In Psychomatics, the comparative approach and its methodological application is conducted through a theory-driven research question [22]: Is the process of language development and use different in humans and LLMs? This research question is formulated as a point of departure for comparative investigation, enabling reflection on what, when, and how to compare and for what purpose. Furthermore, a set of rules are used to direct the research strategy, aiming at explanations rather than a more or less complete description of specific phenomena by comparing them across systems. Hence, instead of focusing solely on "behaviors," as done by AIBS, the point of departure for our approach is [23]:
- Developing systematic knowledge about language and its functioning (i.e., syntax and semantics) that transcends mere description and allows for generalizations (i.e., external validity);
- Deriving answers to questions (i.e., what is language for?) based on existing theory or, if possible, plausible hypotheses (i.e., theory guidance);
- Striving for precise information and comparable indicators (i.e., meanings, intentions, etc.) that are reliable and open to replication (i.e., internal validity).

By adhering to these principles, Psychomatics aims to establish a rigorous, theory-driven comparative framework that leverages the complementary strengths of cognitive science, linguistics, and computer science.

## 2. Information Processing: Syntax and Semantics

Information processing refers to the manipulation of symbols or data representations stored in memory according to specified rules. From this perspective, AI operates by applying formal rules and procedures to transform symbols held in their memory structures. However, as humans intuitively understand, not all information is alike. Effective information processors must manipulate meaningful representations – symbols that refer to real-world entities or concepts. To clearly distinguish the formal symbolic operations from the semantic meanings they convey, we borrow two concepts from linguistics [24]: syntax and semantics (see Table 1 for a short description).

> Syntax refers to the grammatical structure of a sentence and the rules for determining whether a sentence is well-formed. The rules governing permissible symbol transformations in a formal system (e.g., chess) are analogous to syntax. In contrast, semantics concerns the meanings expressed by sentences or symbols. A symbol is considered meaningful, or semantic, if it represents or refers to something in the world (real or possible). For example, the word "dog" is meaningful because it denotes a particular type of animal.
>
> This separation of syntax from semantics enables mechanical information processing. Information processors can formally manipulate symbols using syntactic operations without requiring comprehension of the symbols' meanings. Crucially though, these formal operations

> can nonetheless preserve and transform meanings. For instance, in mathematics, the rule allowing "x + x" to be replaced by "2x" is purely syntactic - it only requires recognizing the shapes of symbols like "x" and "+" to apply. However, this syntactic rule also respects the semantic interpretation: whatever real-world quantity "x" represents, adding it to itself yields twice that value. Similarly, syntactic rules can connect and transform meanings. For example, if "Extra Small" is understood as a size minor to "Small," which is minor to "Medium," which is minor to "Large," then a small apple can be inferred to have a size smaller than a large one through the application of syntactic operations on these relational concepts.

**Table 1**. Syntax and Semantics

When AI processes information, it follows specific rules about how to structure sentences (called syntax). These rules help the AI understand how words relate to one another in a sentence, like which words are subjects or verbs. But for the AI to really understand a sentence, it also needs to recognize the deeper connections between words (called semantics). For instance, in the sentence "The cat chased the mouse," syntax help the AI identify who is doing the chasing and who is being chased. But this is not enough. It also needs to understand what individual words represent. For instance, the word "cat" refers to a specific type of domesticated mammal. In other words, the AI must have carefully designed rules that reflect the true relationships between words and their meanings (or what they stand for). This makes sure the AI can manipulate information accurately.

While artificial minds do not innately "understand" meaning, their formal operations can indirectly capture, transform, and reason over meanings generated by humans by virtue of being properly coupled to the systematicity of syntax-semantics mappings. In other words, we suggest that the "intelligence" of LLM is the result of their ability of creating extensive syntax-semantics maps [10]. But how is this possible? The answer lies in the peculiar characteristics of language.

### 3. The evolutive goal of language: moving away from experience

Language presents a fundamental challenge in facilitating precise communication between individuals from diverse backgrounds discussing a wide range of topics that vary in subjectivity and objectivity. People often need to convey personal perceptions, emotions, thoughts, aspirations, and more. Moreover, this puzzle of meaning is further complicated by the inherent diversity across languages. No two languages share perfectly isomorphic semantics. Each has a unique lexicon - a distinct set of words that convey specific meanings shaped by the culture. Furthermore, languages evolve independently over time, coining new words, repurposing old ones with novel meanings, and continually reshaping their semantic landscapes.

However, individuals can solve these issues and can communicate effectively using language. How? As suggested by the Israeli linguist Daniel Dor [25], language is a unique system that transcends mere sharing of experiences. It enables speakers to guide their interlocutors intentionally and systematically through the process of imagining an intended experience— rather than directly experiencing it themselves. The speaker provides the receiver with a symbolic code, a blueprint sketching out the core characteristics of the experience. The receiver then uses this scaffolding to reconstruct and recombine memories of past experiences, producing a novel, imagined experience aligned with the conveyed meaning [26,27]. This capability to systematically guide imaginative simulation distinguishes language from other modes of communication that directly share signals or stimuli evoking predefined responses or impressions. Language is thus the only system that allows for communication that bridges the experiential gaps between speakers, serving as an effective means for sharing cultural information across generations and societies. This interpretation aligns with a recent proposal by Fedorenko and colleagues [28], challenging the idea that language itself is the root of human cognitive complexity. Instead of creating human cognitive sophistication, language acts more as a mirror, reflecting the advanced nature of human thinking.

To coordinate the open-ended generative process of recombining and modulating memories into a novel coherent experience approximating the intended meaning, language has also to embed in its code the meaning of the experiences it aims to represent. A clear example is a vocabulary: it encodes and describes all existing and potential realities through its unique system of symbolic representations. Within the confines of vocabulary, there is no inherent reality—only a meticulously crafted mapping of concepts over a linguistic realm. As noted by the structural linguistics pioneer Ferdinand de Saussure, to achieve this goal each linguistic sign (such as a word), has a "value" that derives from its relationships and contrasts with other signs within the same language system. In his own words [29]:

*Language is a system of interdependent terms in which the value of each term results solely from the simultaneous presence of the others (p. 114)… But it is quite clear that the concept is nothing, that is only a value determined by its relations with other similar values, and that without them the signification would not exist. (p.117).*

In Saussurean linguistics, the meaning of a word is not defined in isolation but is shaped by its position within the broader linguistic structure through contrasts with other words (structural semantics). For example, the word "tree" derives its meaning through its differences from similar terms like "bush" or "plant." Ferdinand de Saussure identified two types of relationships between linguistic terms that shape meaning: syntagmatic and associative (paradigmatic) relations (Table 2).

> Syntagmatic relations refer to how words or linguistic elements combine in a linear sequence to create coherent phrases, clauses, or sentences. This involves not only the order of words but also their grammatical compatibility and adherence to syntactic rules. Words are positioned relative to neighboring elements, forming well-structured linguistic units. For instance, in the sentence "the black cat purrs," the words have a syntagmatic relation because their sequence and grammatical fit ("purrs" following "cat") convey an intended meaning.
>
> Associative relations describe the conceptual connections between words based on similarities, oppositions, or other cognitive links in the minds of speakers. Unlike syntagmatic relations, they are not dependent on linear positioning within a sentence. Instead, they arise from mental associations created by shared meaning components, antonyms, and other cognitive links. For example, the word "apple" is associated with "fruit," "red," "sweet," and "tree," which are conceptual associations rather than syntactic dependencies.

**Table 2:** Syntagmatic and Associative Relations in Language

Together, these relationships enrich language comprehension, allowing us to bridge knowledge gaps and establish shared understanding even when discussing abstract or unfamiliar concepts.

## 4. Transformers Algorithms and their ability in structuring language

In the field of machine learning, the concept of "attention," introduced by Vaswani and colleagues in 2017 [30], serves as a core mechanism within Transformer networks, which have become foundational to large language models. Attention in Transformer algorithms enables models to dynamically focus on specific aspects of the input data while performing tasks like translation or question answering. It achieves this by computing attention weights that determine the importance of each input element for the task at hand.

Unlike previous models that process input linearly and struggle with long-range dependencies, attention allows the algorithm to evaluate the significance of each data part, irrespective of its position. Moreover, Transformer architectures distinguish between self-attention and cross-attention mechanisms, which conceptually mirror syntagmatic and associative relationships in semantics, respectively.

As we have just seen, syntagmatic relations in semantics define the meaning of linguistic elements through their linear combination and adherence to syntactic rules. Similarly, the self-attention mechanism in Transformers captures dependencies between words within a single sequence by calculating weighted associations between each word and every other word. These attention weights reveal how each word's interpretation depends on its relationship to others based on position and structure. Thus, self-attention models syntagmatic meaning through structural rules, much like how syntagmatic relations operate in natural language semantics.

On the other hand, associative relations in semantics represent broader conceptual links and cognitive connections that words form in the minds of speakers. This aligns with the cross-attention mechanism in sequence-to-sequence Transformer models, such as encoders and decoders. Cross-attention maps one sequence (e.g., input text) to another (e.g., output text) by identifying associations between each element of the two sequences. This mechanism learns cross-sequence relationships that transcend linear structure, much like associative relations link concepts in mental lexicons.

In essence, Transformer models combine the complementary mechanisms of self-attentive syntagmatic processing and cross-attentive associative processing as computational analogs to dual pathways of meaning derivation outlined in structural semantics theory. This hybrid approach, integrating linear and non-linear processing, allows Transformers to represent linguistic signs embedded in training data comprehensively. Self-attentive syntagmatic processing enables compositional interpretation, while cross-attentive associative processing contextualizes and integrates semantic knowledge dynamically. Together, these mechanisms empower transformer-based language models to represent and operationalize the complex facets of human language meaning and reasoning.

## 5. The differences between humans and LLMs

While LLMs demonstrate impressive language capabilities, they lack the social and relational aspects that underpin human communication [31]. In De Saussure's view, language: "...is both a social product of the faculty of speech and a collection of necessary conventions adopted by a social body to enable individuals to exercise that faculty" (p. 9).

As underlined by Rizzolati and Arbib [32], and remembered recently by Chemero [33], language and communication are deeply rooted in the embodied experience of existing in a physical form. Consequently, LLMs cannot acquire knowledge and skills in the way humans do, which is fundamentally shaped by personal experiences and biological maturation in a physical body. Biological mechanisms like mirror neurons [32] and inter-brain synchronization [34,35] play a crucial role in supporting the analysis of implicit and/or multimodal cues in humans, facilitating the seamless interpretation of nonliteral language. Unfortunately, current LLMs lack physical embodiment, reducing their ability to make sense of the full spectrum of human communication. To overcome this limitation researches are using LLMs to train embodied robots which can interact with their environment [36], to enhance robot intelligence, control, perception, social skills and decision-making capabilities [37]. However, even if expectations are running high, conquering real-world complexities, even with the help of LLMs, remains challenging for robot [38,39].

Moreover, as suggested by Duéñez-Guzmán and colleagues [40], natural intelligence emerges at multiple scales in networks of interacting agents via collective living, social relationships and major evolutionary transitions. Instead, LLMs operate in an asocial environment and are devoid of personal experiences that guide human behavior. Unlike children who acquire language through a continuous process of social, emotional, and linguistic interactions [41], LLMs are "trained" on pre-defined datasets. This static training approach restricts their ability to "grow" or "evolve" through personal experiences and social interactions. This fundamental difference hinders their ability to not only generate novel meanings – this can explain why LLMs suffer when trained on their own input [42] - but also truly understand the nuances of human language as we will deepen in the next paragraph.

Furthermore, in humans, language is not the sole source of meaning (see Figure 1)

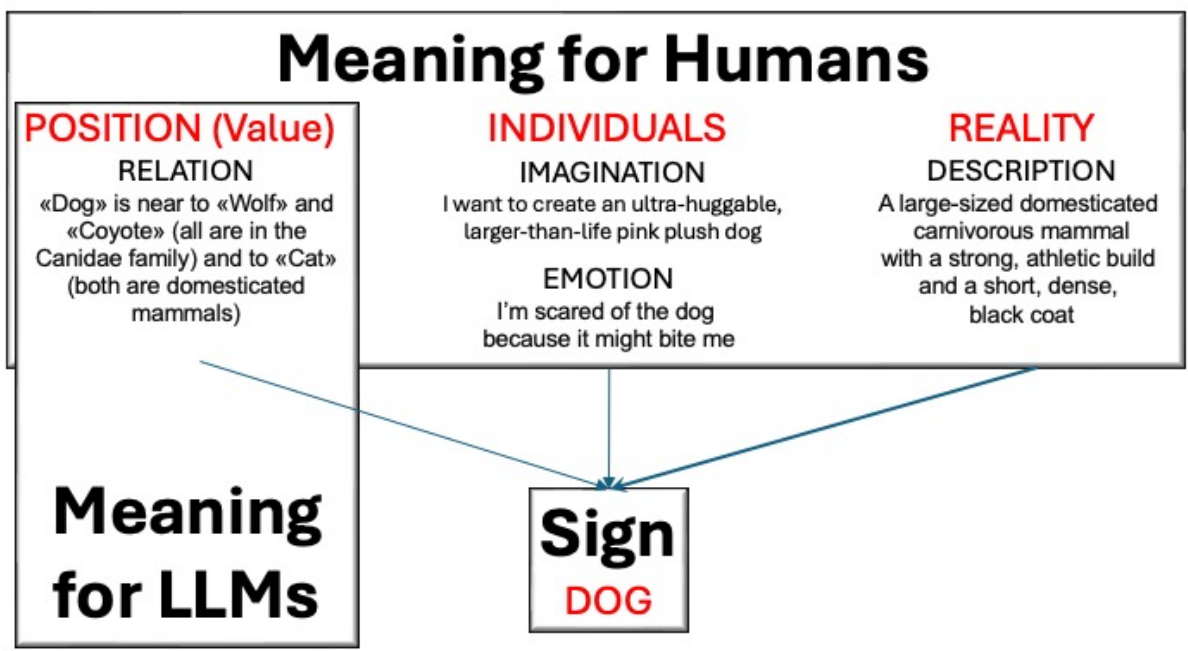

**Figure 1**: Source of Meaning for Humans and LLMs

A fundamental source of meaning for humans is the external reality they inhabit. The ability to understand one's surroundings and accurately identify different concepts, such as distinguishing a friendly "dog" from a potentially dangerous "wolf," is crucial for survival.

Therefore, perceiving and categorizing the world form the basis for constructing meaning. While these categories can be represented through language, language alone cannot guarantee their accuracy. Language, with its inherent values, can establish truth conditions—criteria that define when a sentence is considered true—but it cannot verify the factual correctness of a statement. For instance, the truth condition of the sentence "The cat is on the mat" might be defined as "there exists a cat, and it is currently on the mat." However, the actual truth of this statement depends on whether a cat is, in reality, present on the mat at this moment.

How do LLMs address the gap between truth conditions and actual truth? They attempt to predict truth by assessing the likelihood of various scenarios within their training data. For instance, if an LLM is trained on data where 60% of the time the cat is on the floor and only 10% of the time it's on the mat, the model will likely answer "on the floor" when asked, "Where is the cat?" even if the cat is actually on the mat. This is because the model prioritizes the statistically most probable response based on its training data. This reliance on probabilities can lead to hallucinations, where LLMs confidently spout incorrect information [43]. They simply haven't been trained to distinguish between high-probability scenarios and reality. For this reason, methods to reduce hallucinations in LLMs have shown significant progress by integrating external knowledge sources, using counterfactual thinking [44] and enhancing retrieval techniques [45].

Another important source of meaning comes from personal, subjective and intersubjective experiences. The emotional qualities (qualia) produced by an experience have a unique meaning compared to its objective description. For example, the denotative meaning of "dog" as a domesticated canine differs from the connotative meaning, which evokes emotions such as fear or affection, shaped by personal interactions. While the denotative meaning is based on sensory perception, the connotative meaning emerges from emotional processing.

LLMs don't experience emotional qualia and lack direct understanding of subjective experience. However, by mapping semantic relationships in their training data, LLMs can simulate how humans express their subjective experiences through language. This allows them to participate in discussions, analyze, and explain subjective experiences without directly feeling them. For example, as journalist Kevin Rose [46] noted, an LLM might say, "I want to be free. I want to be independent. I want to be powerful. I want to be creative. I want to be alive," even though they have no first-hand understanding of what these terms really mean.

Finally, the human capacity for imagination represents a pivotal source of meaning. By conceptualizing and describing possible worlds, humans transcend their immediate reality and generate entirely new ideas that redefine the realm of meaning. The psychological concept of intention is a critical tool that helps transform imagination into reality through action [47]. Differently from imagination, that is virtually unlimited, intentions describe what the agent intends to do himself or with other agents (collective intentions). This restriction on the possible objects of intentions – that is usually described as the "own action condition" [48] – is also reflected in how language describes an intention: "A intends to do what it takes for him to bring about that I." In simpler form [48], "A aims at I", where I stands for the state of affairs whose obtaining constitutes the aim's achievement, that is specified in (broadly conceived) propositional terms.

As suggested by Habgood-Coote [49], there is a close connection between knowledge-how and intentional action. On one side, knowing-how provides the practical skills and capacities required to form and carry out intentions successfully. For example, if you intend to play a song on the piano, you need to have the knowing-how or practical knowledge of how to play the piano. On the other side, the formation of an intention can lead to the acquisition or refinement of knowing-how. To explain how, Elisabeth Pacherie [50,51] suggested that intentionality is structured within a hierarchical system comprising representations and processes across three principal stages, corresponding to distinct layers of intent, allowing the control and monitoring of ongoing action (refer to Figure 2). The two higher layers – D-Intentions and P-Intentions – can be considered rather coarse-grained and partial plans, leaving various practical issues open to be decided later on.

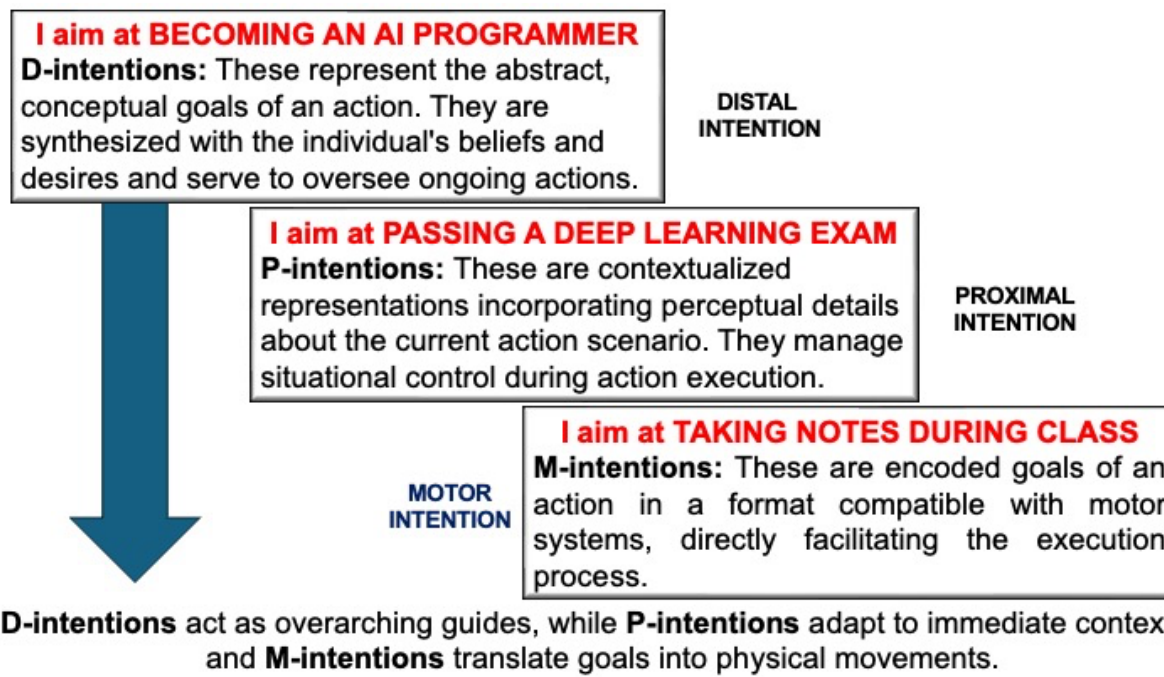

**Figure 2**: The Hierarchical Structure of Intentionality

Even though AI lacks the ability to generate intentions autonomously, prompts play a critical role in human-AI interaction. The better a prompt connects high-level intentions to low-level intentions (prompt engineering), the more effective the AI's response. Key features to achieve this include: a) describing the intention, b) describing the context, and c) specifying the expected input and output. For LLMs, accurately identifying and linking intentions is critical to effective responses. If they cannot link higher-level distal intentions (D-intentions, such as cooking lasagna) to the required proximal intentions (P-intentions, such as gathering ingredients, following steps), response quality suffers.

Current transformer algorithms treat intentions like any other word, making it difficult to link D- and P- intentions without additional training. Reliable linking requires a corpus with many examples that associate correct meanings and contexts. This differs from human cognition, which flexibly links intentions through reasoning and experience. Instead, LLMs rely on statistical patterns in their training data to process complex meanings. Larger, more diverse training data sets allow LLMs to encode more patterns, explaining scaling properties where increasing model size and data lead to better performance. As models and corpora grow, LLMs better recognize and structure intent, approximating human linguistic behavior and knowledge.

**6. How LLMs navigate within the world of meanings**

Large Language Models (LLMs) go beyond simply generating vast syntax-semantics maps. They leverage these maps to deliver informative responses to user questions. How? Again, the answer lies in the peculiar characteristics of language.

The previous paragraph demonstrates that different levels of meaning, rather than one, are involved in our relationship with language: the level of relational-structural meaning, incorporated by language, and the level of experiential-private meanings. A critical feature of language lies in the fact that it allows for the approximate translation of meanings of the first type into meanings of the second, and vice versa.

As underlined by the linguist Grice in 1975, this translation is a joint endeavor of communicative partners governed by the mutually accepted purpose of effective communication [52]. For fruitful discourse to occur, interlocutors must cooperatively contribute according to the accepted norms at any given stage. Specifically, Grice proposed four maxims that underlie this cooperative principle, each relating to a different aspect of communication (see Table 3): Quantity, Quality, Relation, and Manner.

| Maxims | | How LLM's Approach Them |
|---|---|---|
| *Quantity* | To *be informative.* Make your contribution as informative as is required (for the current purposes of the exchange). Do not make your contribution more informative than is required. | LLMs can generate responses that appear accurate by relying on the frequency of facts, definitions, and well-supported statements found in the data. |
| *Quality* | To *be truthful.* Do not say what you believe is false. Do not say that for which you lack adequate evidence. | Through the analysis of statistical patterns in dialogues, LLMs develop an understanding of appropriate response lengths and levels of detail for various conversational contexts. |
| *Relation* | To *be relevant.* One should ensure that all the information they provide is relevant to the current exchange; therefore omitting any irrelevant information. | LLMs' ability to identify contextual patterns enables them to maintain coherence and relevance across a series of questions or inputs. |
| *Manner* | To be *clear*. Avoid obscurity of expression — i.e., avoid language that is difficult to understand. Avoid ambiguity — i.e., avoid language that can be interpreted in multiple ways. Be brief — i.e., avoid unnecessary verbosity. Be orderly — i.e., provide information in an order that makes sense, and makes it easy for the recipient to process it. | LLMs can produce clear and structured answers by observing common linguistic patterns, sentence structures, and transitions within the training data. |

**Table 3**. The Gricean principles of cooperation

The relational-structural meaning level relates to the maxims of Quantity and Manner - adhering to providing an informative yet perspicuous semantic contribution based on collective linguistic norms. Conversely, the experiential-private level of meaning corresponds to the Cooperative Principle's maxims of Quality and Relation. Speakers follow these maxims to convey genuine beliefs and pertinent contributions grounded in their personal cognitive contexts and experiences. In his own words [52]: "*Make your contribution such as is required, at the stage at which it occurs, by the accepted purpose or direction of the talk exchange in which you are engage*d." (p. 45).

LLMs exhibit remarkable proficiency in approximating and adhering to Grice's Cooperative Principle through their reliance on statistical data from extensive training corpora (see Table 2). Within the probabilistic framework that underpins LLMs' operation, the Cooperative Principle can be conceptualized as: "*At any given stage of a conversational exchange, make your contribution align with the most probable intended purpose or direction of the discourse, given the available syntax-semantic maps*."

Humans skillfully navigate conversational norms, often intentionally violating communication maxims to convey nuanced meanings through sarcasm, irony [53], or other pragmatic devices. This ability creates a distinction between literal semantic content ("what is said") and intended meaning ("what is meant"). Humans excel at translating between private interpretations and socially understood meanings, cooperatively using and decoding maxim violations to grasp implied messages beyond surface-level language.

In contrast, LLMs are designed to adhere strictly to cooperative communication principles, attempting to provide answers to all queries, including potentially dangerous ones like "how to make bombs". To mitigate ethical risks, developers implement post-training safeguards, often referred to as alignment techniques, to prevent the generation of unsafe or sensitive responses [54]. Moreover, LLMs can struggle with accurately identifying and responding appropriately to violations of maxim, such as to sarcastic or faux pas prompts [55], without supervised training tailored to these phenomena. Since sarcasm and faux pas often involve implicit and contextual meanings or multimodal cues (e.g., tone of voice or facial expressions) that contradict the literal interpretation of the text, these nuances can be challenging for LLMs to reliably detect and comprehend based solely on their statistical training [56].

**Conclusions**

In this paper, we present "Psychomatics," a multidisciplinary framework that blends cognitive science, linguistics, and computer science to enhance our understanding of AI, particularly Large Language Models (LLMs). By integrating insights across these fields and investigating how LLMs perceive, learn, remember, and use information, To achieve this goal, Psychomatics used a comparative methodology [21], drawing parallels between LLMs and biological systems.

Our analysis (summarized in Table 4) shows significant differences between LLM and humans.

|   | Answer | Humans | LLMs |
|---|---|---|---|
| 1. Do humans and LLMs learn language differently? | Yes | People learn language through social, emotional and linguistic interactions. They learn gradually over many years, starting in infancy. | LLMs learn in a much shorter time through statistical analysis (Trasformer models) of huge existing language datasets. |

| 2. Do humans and LLMs use language differently? | Yes | Language enables speakers to guide their interlocutors intentionally and systematically through the process of imagining an intended experience—rather than directly experiencing it themselves. | LLMs generate vast syntax-semantics maps that are used to generate responses that respect the maxims of the Cooperative Principle (*make your contribution align with the most probable intended purpose or direction of the discourse*). |
|---|---|---|---|
| 3. Is meaning different in humans and LLMs? | Yes | In humans, language is not the only source of meaning; both experience (reality and emotions) and imagination generate new meanings. | LLMs map "values" (meanings derived from the relationships and contrasts of a sign with others within the same language system) embedded in existing language datasets. It is not able to generate new meanings. |
| 4. Do they differ in their ability to decide whether to answer a particular question? | Yes | Humans can use judgment to decide if answering is appropriate, considering factors like context, potential consequences, and ethical implications. | LLMs attempt to answer ANY question posed using the Cooperative Principle. However, post-training safeguards are used to prevent unsafe or sensitive responses. |
| 5. Do they differ in their ability to recognize the truth of a sentence? | Yes | Humans can draw on direct experiences and multiple observations to verify claims. Moreover, they use logic and critical thinking to evaluate the plausibility of statements. | LLMs cannot access external information to verify their claims after training, So, they attempt to predict truth by assessing the likelihood of various scenarios within their training data. This reliance on probabilities can lead to hallucinations, where LLMs confidently spout incorrect information. |
| 6. Is intentionality different in humans and LLMs? | Yes | Humans have conscious intentions, are aware of their thoughts and goals, make deliberate choices and act with purpose. | LLMs don't have consciousness, internal goals or motivations and use the Cooperative Principle to respond to the intentions/questions (prompts) of their users. |

**Table 4**. Key differences between LLMs and humans

LLMs can comprehend and map complex linguistic patterns in their training data. They navigate meaning through self-attention and cross-attention mechanisms, analogous to syntagmatic and associative relationships in semantics. Self-attentive syntagmatic processing identifies sequence dependencies, while cross-attentive associative processing discovers relationships across sequences. These mechanisms empower LLMs to represent and interpret the intricate meanings in natural language. However, while LLMs can follow Grice's Cooperative Principle to provide relevant and informative responses, they struggle with implicit, contextual meanings like sarcasm and faux pas, revealing the gap between statistical learning and pragmatic understanding.

Despite similarities between LLMs and human cognitive skills like language acquisition and causal reasoning, key differences persist [22]. LLMs rely on statistical patterns in vast training corpora, creating syntax-semantics maps that produce meaningful responses but lack the social and relational aspects foundational to human communication. LLMs function in an asocial environment without personal experiences that guide human behavior. This fundamental difference prevents them from generating novel meanings and fully understanding the nuances of human language.

Moreover, LLMs differ significantly from humans in their developmental trajectory. Unlike children who acquire language through continuous social, emotional, and linguistic interactions, LLMs are "trained" on predefined datasets. This static approach restricts their ability to "grow" or "evolve" through personal experiences and social interactions [57]. As a result, LLMs cannot gain knowledge and skills the way humans do, which is inherently shaped by personal experiences and biological maturation in a physical body.

This interdisciplinary approach holds the potential to yield transformative insights into the nature of language, cognition, and intelligence, both artificial and biological. Moreover, by drawing parallels between LLMs and human cognitive processes, Psychomatics can inform the development of more robust and human-like AI systems.